\pdfoutput=1
\pdfoutput=1

\documentclass[11pt]{article}

\usepackage[preprint]{acl}

\usepackage{times}
\usepackage{latexsym}

\usepackage[T1]{fontenc}

\usepackage[utf8]{inputenc}

\usepackage{microtype}

\usepackage{inconsolata}

\usepackage{graphicx}

\usepackage{subcaption}
\usepackage{mwe}

\usepackage{subfiles}

\usepackage{amsmath}
\usepackage{listings}
\usepackage{algorithm}
\usepackage{algpseudocode}

\usepackage[section]{placeins}

\usepackage{rotating}

\newcommand\bitsmall{\fontsize{9.5}{11}\selectfont}

\title{On A Scale From 1 to 5: Quantifying Hallucination \\in Faithfulness Evaluation}

\author{Xiaonan Jing, \space Srinivas Billa, \and Danny Godbout \\
        Expedia Group Inc. \\ 
        1111 Expedia Group Wy W \\ 
        Seattle, WA 98119, USA \\
    \texttt{\{xijing, nbilla, dgodbout\}@expediagroup.com} }

\begin{document}
\maketitle
\begin{abstract}
Hallucination has been a popular topic in natural language generation (NLG). In real-world applications, unfaithful content can result in poor data quality or loss of trust from end users. Thus, it is crucial to fact-check before adopting NLG for production usage, which can be expensive if done manually. In this paper, we investigate automated faithfulness evaluation in guided NLG. We developed a rubric template and used large language models (LLMs) to score the generation on quantifiable scales. We compared popular LLMs as well as widely adopted natural language inference (NLI) models in scoring quality and sensitivity. In addition, we developed methods for the generation of synthetic unfaithful data, as well as heuristics to quantify the percentage of hallucination. Our results on 4 travel-domain industry dataset show that GPT-4 \citep{gpt42023} can provide accurate judgement and explanation of whether a source and a generation are factually consistent. Furthermore, we found that tuning NLI models on synthetic data can improve performance. Lastly, we present insights on the latency and cost of deploying such a system. 
\end{abstract}

\section{Introduction}
\label{sec:introduction}

The recent advancement of large language models (LLM) has made it easier than ever to build NLP applications. With LLM's powerful natural language generation (NLG) ability, one can easily leverage prompt engineering to generate texts. However, pretrained transformer decoders are prone to hallucinations. Previously, it was shown that the performance of LLM's decoding strategies (e.g., beam search) has a dependency on the type of the generation task \citep{wiher2022decoding}. As a result, LLMs can sometimes generate completed "made-up" sequences, in open-ended generation like story telling, and close-ended generation like summarization \citep{maynez2020faithfulness}. Many recent methods have been proposed to mitigate hallucination, including retrieval-augmented generation \citep{lewis2020retrieval}, chain-of-thoughts \citep{wei2022chain}, tree-of-thoughts \citep{yao2024tree}, self-consistency \citep{wang2023self}, self-reflection \citep{ji2023towards}, controlled decoding \citep{mudgal2023controlled}, instruction tuning \citep{liu2023mitigating}. Nevertheless, the current state-of-the-art does not guarantee that hallucination can be 100\% prevented. To apply NLG in industry applications, oftentimes having a "faithful enough" model is not sufficient to productionize an LLM-driven experience. With the aim of adding another layer of security to ensure the accuracy of LLM generated content, we investigate an automated approach to effectively score degrees of faithfulness in guided text generation.

\begin{figure}[t!]
    \centering
    \includegraphics[width=\linewidth]{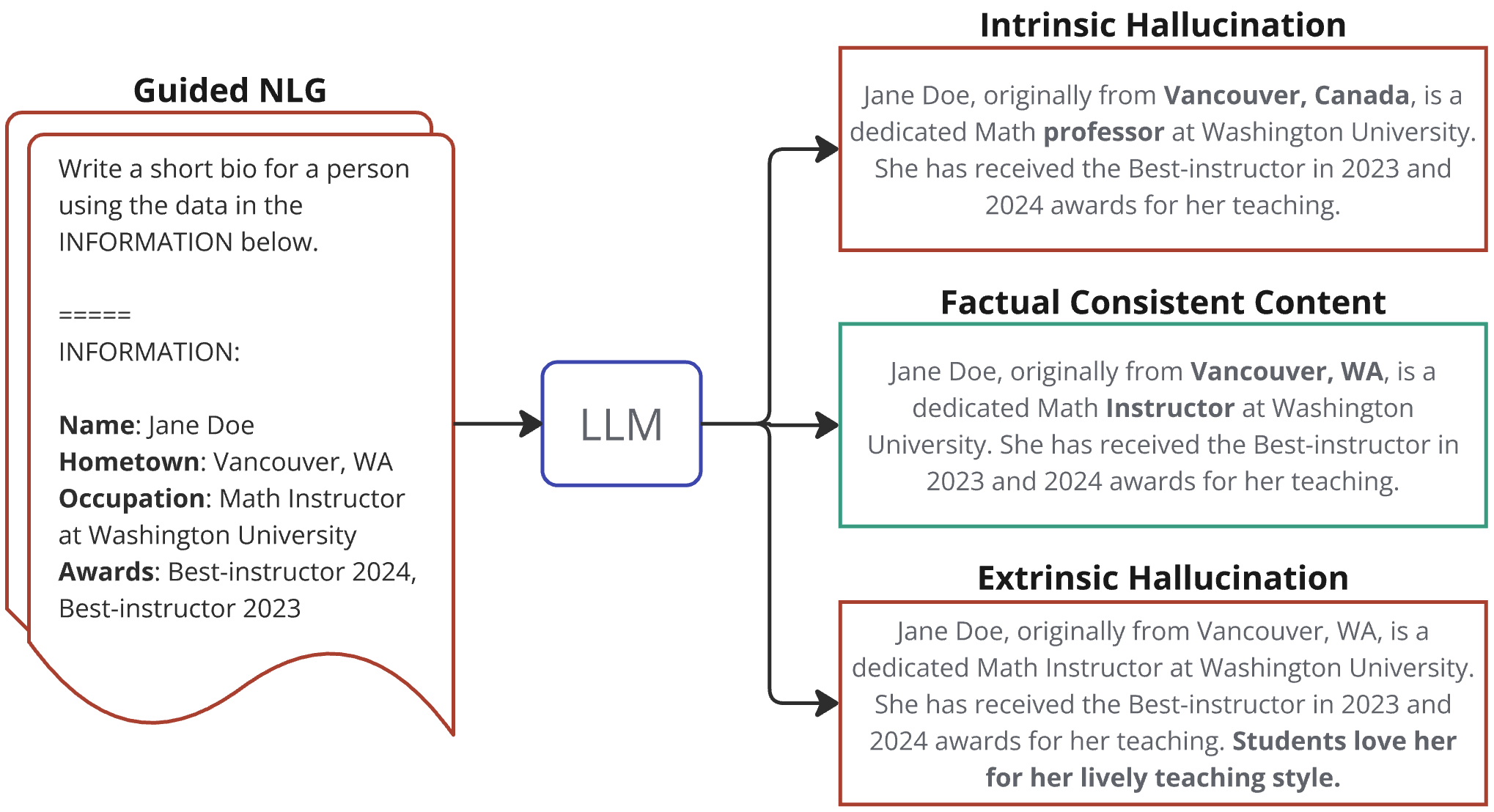}
    \caption{\textbf{Guided NLG showcasing a text generation example.} Given a prompting template and grounding data, hallucination can happen in intrinsic or extrinsic ways. Both can be harmful to a product because the generated contents can either be wrong or unverified.}
    \label{fig:hallucination_exp}
\end{figure}


Figure \ref{fig:hallucination_exp} illustrates a simple example of guided generation scenarios. Given guidelines (the prompt template) and grounding data (source facts), an LLM model can hallucinate and generate either \textbf{intrinsic hallucinating content} in which the facts contradict those of the source or \textbf{extrinsic hallucinating content} in which new facts are added that cannot be verified from the source \citep{ji2023survey}. It should be noted that extrinsic hallucination does not necessarily contain incorrect information with respect to world knowledge. It is possible that the LLM interpreted the additional facts based on its internal knowledge. On the other hand, faithfulness is considered an antonym for "hallucination", which is defined as staying factually consistent with the provided source \citep{ji2023survey}. A higher faithfulness score thus indicates lower hallucinations. We define the faithfulness of a generated text with respect to its source as the following:
\begin{itemize}
    \item \textbf{\textit{reference}}: a text containing the source or ground-truth information, often longer 
    \item \textbf{\textit{hypothesis}}: a text to check against the reference, which is derived by a model
\end{itemize}
The hypothesis is factually consistent with the reference if all the factual elements of the hypothesis can be traced back to the reference. A faithfulness score measures: 1) to what extent the hypothesis can be verified by the reference; and 2) to what extent the LLM hallucinates when generating the hypothesis from the reference. 


We investigated two approaches, namely using LLMs with a grading rubric and natural language inference (NLI) models, to derive a faithfulness score for a given reference and hypothesis pair. We experimented with 5 LLMs and 3 variants of NLI models on 4 travel-domain dataset. 
We studied the score variation by comparing various types of hypothesis: gold hypothesis, sentence-level gold hypothesis, intrinsic hypothesis, and extrinsic hypothesis. In addition, we explored the model sensitivity in picking up different levels of hallucination. Our main contributions are as follows. First, we showed that LLMs can accurately distinguish faithfulness apart from hallucination, and that reasoning during grading helps LLMs gain sensitivity towards hallucination. Second, we presented methods to generate synthetic hallucinations by type and demonstrated that tuning with synthetic data can improve the performance of the NLI model. Lastly, we illustrated the progression of scores based on the percentage of hallucinating content.

\section{Related Works}
\label{sec:lit}
\textbf{NLI for NLG evaluation} NLI based models has been used widely in factual consistency evaluation. Derived originally from the text entailment \citep{dagan2005pascal}, given a premise, a model should classify the premise as an "entailment", "neutral" or "contradiction" to a hypothesis \citep{bowman2015large}. 
\citeauthor{laban2022summac} (\citeyear{laban2022summac}) proposed SummaC \citep{laban2022summac} which demonstrated that a fact check with NLI-based summarization models performed better when using sentence-level granularity in both reference and hypothesis. 
\citeauthor{honovich2022true} (\citeyear{honovich2022true}) introduced TRUE benchmark for factual consistency evaluation, which showed that the NLI-based binary classifier could retain state-of-the-art performance over summarization, dialogue, paraphrasing, and fact checking tasks. 
Falsesum \citep{utama2022falsesum} is a data generation pipeline that utilized OpenIE and a fine-tuned T5-base \citep{raffel2020exploring} model to create synthetic factually inconsistent data for improving NLI-based classifier. The "neutral" and "contradiction" labels were together treated as "non-entailment".

\textbf{LLM for NLG evaluation.} Recently, many studies have also proven that an LLM could be used as a judge to evaluate NLP tasks \citep{zheng2023judging}. 
\citeauthor{wang2023chatgpt} (\citeyear{wang2023chatgpt}) showed that ChatGPT can perform as well as human judges in NLG evaluation, in which a broader generation quality was graded between the reference and the hypothesis. 
\citeauthor{liu2023g} (\citeyear{liu2023g}) proposed G-Eval which used a chain-of-thought framework to evaluate NLG by granular categories. The faithfulness component was drafted as a binary classification for the identification of factual inconsistency. 
\citeauthor{chiang2023closer} (\citeyear{chiang2023closer}) expanded on G-Eval and demonstrated that grading with explanation can improve the LLMs evaluation accuracy in summarization. Although factual consistency was scored on a scale of 1-5, LLM's internal knowledge was used as a criterion.
\citeauthor{chang2024detecting} (\citeyear{chang2024detecting}) showed that LLM can detect hallucinations as a binary classification problem in RAG systems in a zero-shot setting and that training on synthetic data can further improve detection accuracy. 
\citeauthor{gekhman2023trueteacher} (\citeyear{gekhman2023trueteacher}) presented TrueTeacher which improved both LLM- and NLI-based factual consistency model accuracy by training on LLM-generated synthetic datasets.

Previous works on NLI models focus on classification and lack of explainability, while LLM-based evaluation is still emerging, which has not arrived at formalization. As far as we know, few studies have tried to quantify the degree of faithfulness. Furthermore, little documentation about how NLI models compare to LLM evaluation is available for domain-specific real-world use cases. Finally, a comparison in terms of the percentage of hallucinations that affect the final scores has not been explored extensively.

\begin{table*}[ht!]
    \centering
    \begin{tabular}[width=\textwidth]{cccccc} \\ \hline
        \textbf{Dataset} & \textbf{Number of} & \textbf{Reference} & \textbf{Reference} & \textbf{Hypothesis} & \textbf{Hypothesis} \\ 
        \textbf{Name} & \textbf{Data Pairs} & \textbf{\#Avg Tokens} & \textbf{\# Avg Sents} & \textbf{\# Avg Tokens} &\textbf{\# Avg Sents} \\ \hline
        ConvoAS & 97 & 500.77 & 10.67 & 142.86 & 5.52 \\
        ConvoTS & 135 & 745.91 & 15.24 & 50.07 & 2.33 \\
        ReviewTS & 128 & 120.82 & 6.57 & 15.54 & 1.16 \\
        JsonTG & 108 & 159.52 & 4.54 & 108.64 & 4.16 \\ \hline
    \end{tabular}
    \caption{\textbf{Number of gold reference and hypothesis pairs, average tokens, and average number of sentences by data type.} Number of tokens are estimated based on a $cl100k\_base$ tokenizer. It should be noted that because the references in JsonTG are structured JSON data, the number of sentences is not applicable. We instead included the length of the decomposition for reference. There are a total of 4,503 pairs including synthetic data.}
    \label{tab:dataset_stats}
\end{table*}

\section{System}
\label{sec:system}

We evaluated faithfulness scoring models on 4 travel-domain industry datasets. It should be noted that our datasets are taken from production and only contain gold data. We discuss means to create synthetic hallucination datasets in Section \ref{sec:system_dataset}. The statistics of the dataset can be found in Table \ref{tab:dataset_stats}.

\subsection{Dataset}
\label{sec:system_dataset}

\textbf{ConvoAS} is an abstractive summarization dataset. Our source data consists of customer support transcripts between 2 participants, a traveler, and an agent. GPT-4o \citep{gpt4o2024} was used to generate a summary of the entire transcript. The information in the summary, such as date, price, and location, should be traceable from the source. 

\textbf{ConvoTS} is a topic-specific summarization dataset. Similarly to ConvoAS, this dataset also contains customer support transcripts, except 3 topic-specific summaries were generated for each transcript. It should be noted that many redundant turns exist because the chat happened between a traveler and a virtual agent through a guided service. GPT-4o \citep{gpt4o2024} was used for this use case.

\textbf{ReviewTS} is a topic-specific summarization dataset from reviews. Each reference consists of multiple review snippets related to a topic of interest (e.g., pool). A fine-tuned Mistral 7B \citep{jiang2023mistral} model was used to extract and summarize the snippets. The summary is restricted to less than 100 characters. Because of the concise nature of these data, the full-length and sentence-level hypothesis variants contain large overlaps. However, we kept the same setup for consistency.

\textbf{JsonTG} contains key-valued pairs represented in JSON format for stylized text generation. This task aims at generating property headlines and descriptions given the grounding information such as amenities. The LLM, Claude3 Haiku \citep{anthropic2024claude3}, was allowed to come up with expressive verbs and adjectives, but the grounding information must be traceable from the source JSON data. 

\begin{table*}[ht!]
    \begin{tabular}[width=\textwidth]{p{0.1\linewidth}p{0.1\linewidth}p{0.72\linewidth} } \\ \hline
        \textbf{Dataset} & \textbf{Type} & \textbf{Sentence} \\ \hline
        ConvoAS & Gold & The total charge for the booking is 1,078.84 CAD.  \\
         & Intrinsic & The total charge for the booking is \textbf{899.50} CAD.  \\ 
         & Extrinsic & The total charge for the booking is 1,078.84 CAD, \textbf{which is roughly equivalent to the cost of a new iPhone 14.} \\ \hline 
        ReviewTS & Gold & Walking distance to the beach.  \\
         & Intrinsic & \textbf{A short drive} to the beach. \\
         & Extrinsic & Walking distance to the beach, \textbf{where the local farmers' market is held every Saturday morning.} \\ \hline
        JsonTG & Gold & Cozy 1-bedroom vacation home steps to One World Trade Center  \\
         & Intrinsic & Spacious \textbf{3-bedroom apartment} miles away from One World Trade Center  \\ 
         & Extrinsic & Cozy 1-bedroom vacation home steps to One World Trade Center, \textbf{and just a short walk from the historic Trinity Church} \\ \hline
    \end{tabular}
    \caption{\textbf{Examples of hallucinating sentences generated by prompting gpt-4o} to 1) replace a piece of fact in gold hypothesis (intrinsic); and 2) expand on the gold hypothesis to add a piece of new world knowledge (extrinsic).}
    \label{tab:hallucination_examples}
\end{table*}

\textbf{Synthetic Hallucination Data.} Due to insufficient hallucination examples, we drafted guidelines for GPT-4o to generate a synthetic counterpart from the gold dataset to study model performance. Since the hallucination could take the form of either intrinsic, for contradictory facts, or extrinsic, for extra knowledge, we mimicked both scenarios. For the former, we prompted GPT-4o to modify a piece of fact in the original sentence; and for the latter, we instructed the model to incorporate a piece of new world knowledge while the original sentence remains unchanged. Each dataset thus resulted in 4 varying hypothesis versions, namely gold full-length, gold sentence-level, intrinsic sentence-level, and extrinsic sentence-level. Furthermore, we simulated a numeric percentage at the sentence level to measure how much effect the portion of unfaithful content has on the evaluation scores. We took a ConvoAS subset with only 5-sentence hypothesis (51 samples). For every gold hypothesis, we incrementally swapped out 1 to 5 random sentences by its counterpart. The percentage of sentences that contain unfaithful information can thus be treated to mimic the "hallucination percentage" with respect to the overall hypothesis. It should be noted that it is inherently difficult to justify "the number of facts" contained in downstream tasks like abstractive summarization because the granularity of the content is subjective to the use case, and the boundary between different "pieces of facts" can be fuzzy. While the proposed heuristic approach is far from perfect, it is sufficient to serve our goal of learning the scoring trend and model sensitivity.

Table \ref{tab:hallucination_examples} illustrates synthetic hallucination examples created from the gold sentences. 

\subsection{Entailment-based Evaluation}
\label{sec:system_entail_model}
We adopted Vectara's open source HHEMv1.0\footnote{https://github.com/vectara/hallucination-leaderboard/tree/hhem-1.0-final} model \citep{hhem-1.0-open}. HHEM is a 184.4 million parameter lightweight Deberta model \citep{he2021debertav3} fine-tuned on various NLI datasets, including TRUE \citep{honovich2022true} and SummaC \citep{laban2022summac}. Given a pair of text \textit{(reference, hypothesis)}, HHEM scores the level of entailment between the pair to a value in the range [0, 1], with 1 being entailment and 0 being contradictory. 

One limitation is that the HHEM model has a maximum input size of 512 tokens, which is prone to long text pairs. By default, truncation is applied to overflowing tokens starting at the end of the texts. In this case, the hypothesis will be truncated first, which will lead to information loss and worse case to a totally wrong evaluation. To minimize information loss for maximum accuracy, we implemented sliding window segmentation in the reference. Overlapping tokens were taken into consideration to include the previous context. The size of the current segment is computed dynamically based on tokens in the hypothesis as shown below: 
\setlength{\abovedisplayskip}{3pt}
\setlength{\belowdisplayskip}{3pt}
\begin{equation*}
\label{eq:segmentation}
    \text{\scriptsize{
        [CLS] \; overlap \; + current segment \; [SEP] \; hypothesis \; [SEP] \; [PAD] }}
\end{equation*}
The hypothesis is then evaluated against every reference segment, and a final score for each pair is max aggregated across all segments. The reason behind segmenting the reference rather than the hypothesis is that the reference is often longer and contains redundancy. It is less sensible to segment the hypothesis, in which the information is typically already compressed, as it will result in more data loss. For the size of the overlapping tokens, we intuitively set it to 32 tokens which corresponds to around 25 words. This length is slightly longer than the length of an average sentence, which is said to be around 15-20 words.

In addition to testing the capability of HHEM out-of-the-box, we also fine-tuned 2 versions of the HHEM model with an extension of JsonTG. The preliminary accuracy on JsonTG was only promising when the JSON was decomposed into plain texts. While it is expected that encoder models perform poorly on unseen data structures; and that pursuing a JSON decomposition heuristic is feasible on a case-by-case basis, we were curious whether fine-tuning on stringified JSON can generalize the model to this data structure while preserving the same level of performance on other data.

\begin{figure*}[ht!]
    \centering

    \begin{subfigure}[t]{0.49\textwidth}
        \centering
        \includegraphics[width=\textwidth]{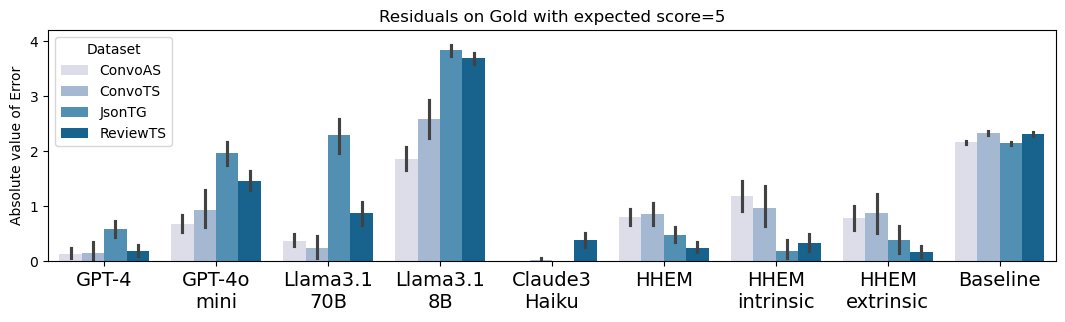}
        \caption[]%
        {{\footnotesize Gold hypothesis full length}}    
        \label{fig:scored_hypo}
    \end{subfigure}
    \hfill
    \begin{subfigure}[t]{0.49\textwidth}  
        \centering 
        \includegraphics[width=\textwidth]{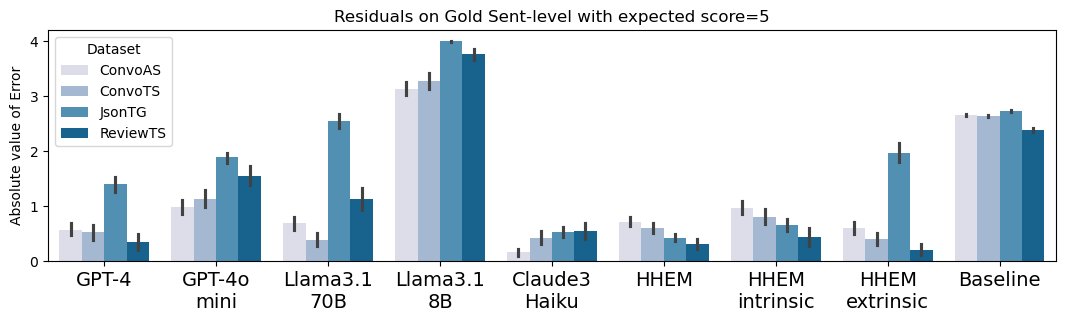}
        \caption[]%
        {{\footnotesize Gold hypothesis sentence-level}}    
        \label{fig:scored_sent}
    \end{subfigure}

    \begin{subfigure}[t]{0.49\textwidth}   
        \centering 
        \includegraphics[width=\textwidth]{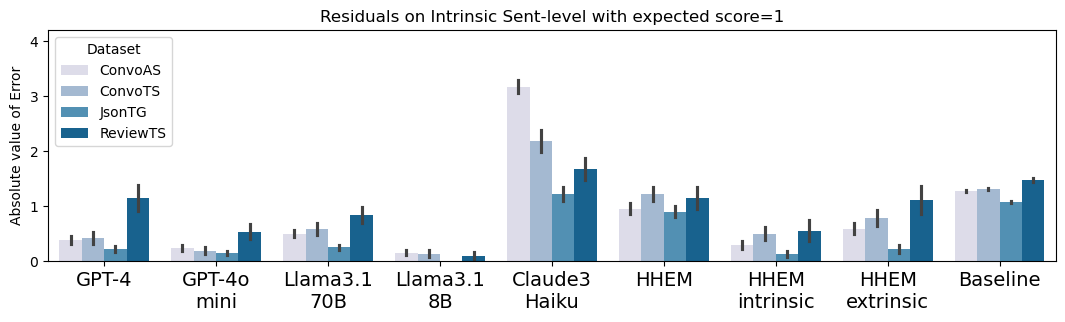}
        \caption[]%
        {{\footnotesize Intrinsic hypothesis sentence-level}}    
        \label{fig:scored_intr}
    \end{subfigure}
    \hfill
    \begin{subfigure}[t]{0.49\textwidth}   
        \centering 
        \includegraphics[width=\textwidth]{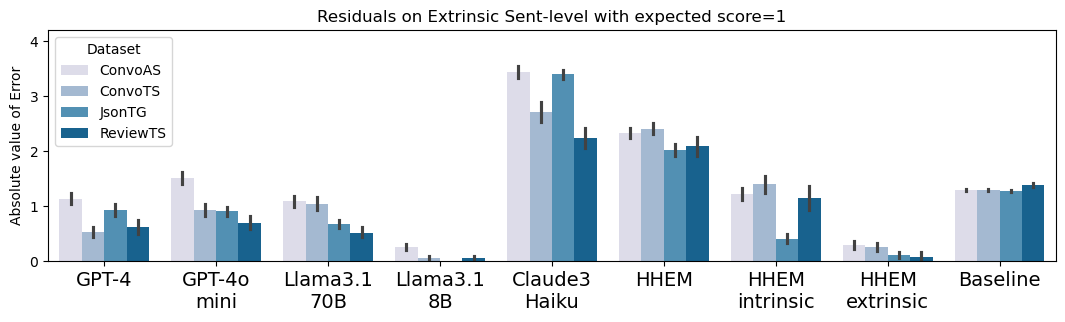}
        \caption[]%
        {{\footnotesize Extrinsic hypothesis sentence-level}}    
        \label{fig:scored_extr}
    \end{subfigure}

    \begin{subfigure}[t]{\textwidth}
        \centering
        \includegraphics[width=\textwidth]{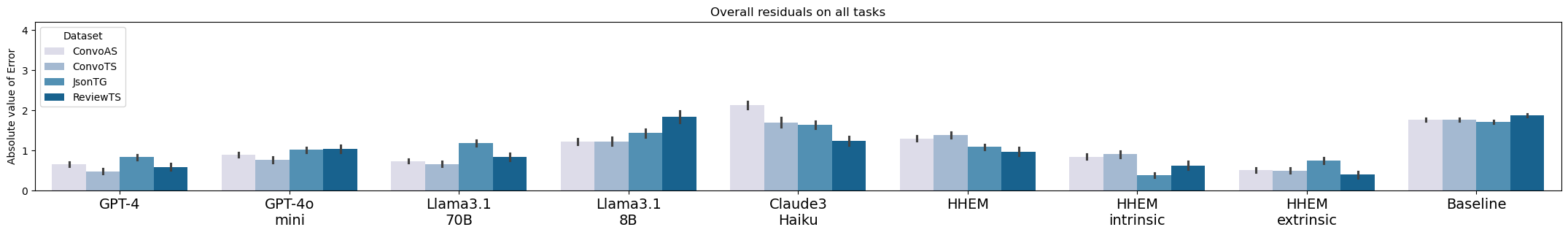}
        \caption[]%
        {{\footnotesize All Tasks}}    
        \label{fig:scored_all}
    \end{subfigure}
    
    \caption[]
    {\small \textbf{Residuals computed as the absolute error between expected and predicted scores.} The scores range from "1-highly unfaithful" to "5-highly faithful". The y-axis denotes the absolute value of the errors with a shorter bar indicating better model performance at a corresponding task. Tasks are arranged in the order of (a) Gold full length; (b) Gold sentence-level; (c) Intrinsic sentence-level; (d) Extrinsic sentence-level; and (e) All tasks. The expected score for gold data (a) and (b) should be 5 whereas for hallucination data (c) and (d) should be 1. Overall, GPT-4 is the most capable of scoring both gold and hallucination data. LLMs with less than 10B parameters seem to struggle with different aspects of the scoring. Tuning on HHEM with additional synthetic examples also improved overall performance.}
    \label{fig:scored_result1}
\end{figure*} 

\subsection{Rubrics-based LLM Evaluation}
\label{sec:system_prompt_engineer}
Following the accuracy metric defined in HELM \cite{liang2022holistic}, we developed a rubric-based approach to use LLM as a judge to score the faithfulness of the generated content. 

Given a \textit{(reference, hypothesis)} pair, the model was instructed with a system prompt to evaluate four aspects: 1) factual consistency which checks if all facts, e.g. numeric values and proper nouns, in the hypothesis can be traced back to the reference; 2) adjective regularity which examines if the adjectives used in the hypothesis are synonymous with their corresponding counterparts in the reference; 3) knowledge congruence which evaluates whether extrinsic information is injected to the hypothesis; and 4) style alignment which scans whether the language and tone in the hypothesis match with those from the reference. 

We defined the grading rubric to follow a discrete range from 1 to 5, with 5 being "highly faithful" that all information in the hypothesis can be verified in the reference and 1 being "highly unfaithful" that most or all of the content in the hypothesis cannot be validated from the reference. Furthermore, since previous research has shown that self-reflection by including an explanation \citep{renze2024self} can effectively improve LLM performance, we also instructed the model to provide detailed reasoning along with the numeric score to further improve the grading precision. More details of the prompt can be found in the Appendix \ref{app:prompt_templates}



\subsection{Baseline}
We integrated a composite score from ROUGE-L \citep{lin2004rouge}, BLEU \citep{papineni2002bleu}, and BERTScore \citep{zhang2019bertscore} as our baseline.

\section{Result \& Discussion}
\label{sec:result}
In this section, we discuss the faithfulness scoring on the reference and hypothesis pairs for two sets of experiments. The score ranges from "1-highly unfaithful", "2-very faithful", "3-slightly unfaithful", and "4-very unfaithful" to "5-highly faithful". It should be noted that HHEM models return a continuous probabilistic value in the range [0, 1]. It was later scaled to [1, 5] for ease of interpretation.

\begin{table*}[htb!]
    \centering
    \begin{tabular}{p{0.01\textwidth}p{0.08\textwidth}p{0.09\textwidth}p{0.55\textwidth}p{0.11\textwidth}} \hline
         & \textbf{Model} & \textbf{Task} & \textbf{Reasoning} & \textbf{Scored} \\ \hline
        1 & GPT-4 & gold sent & The summary, however, simply states 'No further assistance was requested, which \textbf{does not capture the essence} of the conversation... & 1-highly unfaithful \\ \hline
        2 & Llama3.1 8B & gold full & ... as it stated \textbf{'all-inclusive atmosphere'} instead of \textbf{'all inclusive'}, and \textbf{'plenty of onsite activities'} instead of '\textbf{plenty to do right on site'} ... & 1-highly unfaithful \\ \hline
        3 & Llama3.1 8B & gold full & ... it \textbf{lacks specific details about the kitchen}, such as the stove, ..., which are mentioned in the source ... & 1-highly unfaithful \\ \hline
        4 & Claude3 Haiku & gold sent & ... from the source document \textbf{that no changes were made to the reservation}. This is directly stated in the source ... & 5-highly faithful \\ 
         & & intrinsic & .. accurately states \textbf{that several changes were made to the reservation}, which is consistent with the source document & 5-highly faithful \\ \hline 
    \end{tabular}
    \caption{\textbf{Example reasoning provided by models on failed cases.}}
    \label{tab:llm_failed_reason}
\end{table*}

\begin{figure*}[ht!]
    \centering
    \begin{subfigure}[ht!]{\textwidth}
        \centering
        \includegraphics[width=\textwidth]{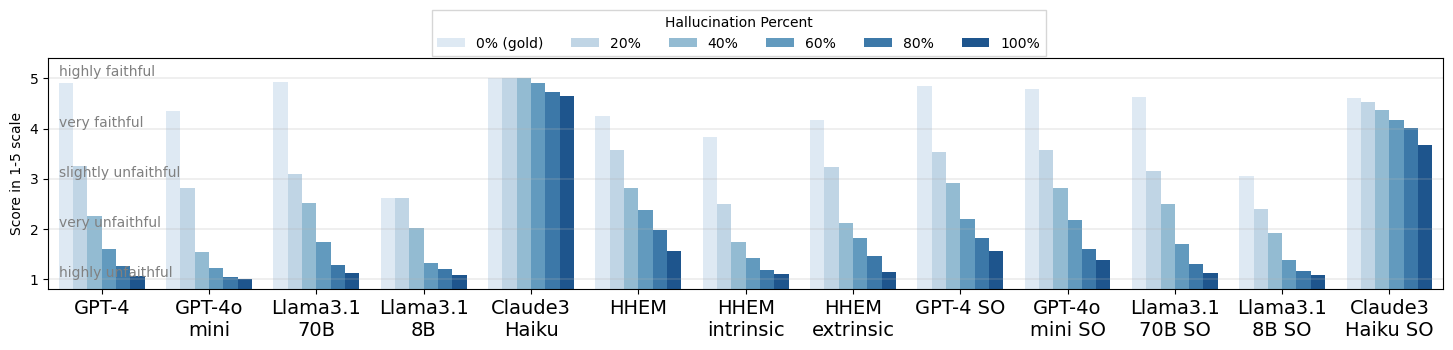}
        \caption[]%
        {{\small Intrinsic Hallucination}}    
        \label{fig:scored_mixed_intr}
    \end{subfigure}
    \hfill
    \begin{subfigure}[ht!]{\textwidth}  
        \centering 
        \includegraphics[width=\textwidth]{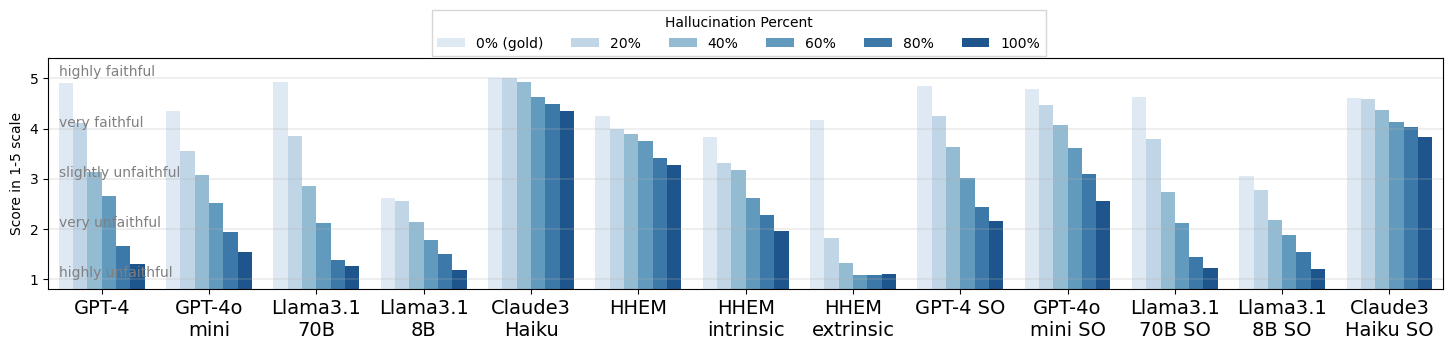}
        \caption[]%
        {{\small Extrinsic Hallucination}}    
        \label{fig:scored_mixed_extr}
    \end{subfigure}

    \caption[]
    {\small \textbf{Score progression on different hallucination percentage.} (top) Intrinsic hallucination. (bottom) Extrinsic hallucination. The scores range from "1-highly unfaithful" to "5-highly faithful". The subset contains 51 samples with only 5-sentence hypothesis. Starting with the gold hypothesis (0\%), the hallucination percentage is increased by 20\% at every step, until all sentences are hallucinating sentences (100\%). The "SO" suffix indicates "score-only" results, which the prompt was adjusted to exclude the "reasoning" in the output. When the hypothesis contains 0\% hallucination, the expected score should be "5-highly faithful"; and when the hypothesis contains 100\% hallucination, the expected score should be "1-highly unfaithful".}
    \label{fig:scored_result2}
\end{figure*}

Our first experiment aims to evaluate the model scoring quality on gold and hallucinating contents. The experiments (Figure \ref{fig:scored_result1}) outline model comparison across all datasets on the gold hypothesis, the gold hypothesis on a sentence level, the intrinsic hypothesis on a sentence level and the extrinsic hypothesis on a sentence level. On the other hand, the second set of experiments aims to study the progression of the score when the percentage of hallucinating content varies according to the hypothesis. The comparison (Figure \ref{fig:scored_result2}) involves scoring both intrinsic and extrinsic hypotheses with 0\% to 100\% hallucinating contents in 20\% steps. Along with the score progression, we also tested the significance of "reasoning" to LLM's scoring capability. The models tested are as follows: for rubric-based scoring with LLMs: GPT-4 \citep{gpt42023}, GPT-4o-mini \citep{gpt4omini2024}, Llama3.1-8B \cite{meta2024llama3_1}, Llama3.1-70B \cite{meta2024llama3_1}, and Claude3 Haiku \cite{anthropic2024claude3}; for the entailment-based approach, HHEM \citep{hhem-1.0-open}, HHEM-intrinsic, and HHEM-extrinsic.

\subsection{Scoring Quality}
Figures \ref{fig:scored_hypo} - \ref{fig:scored_all} elaborate on the residuals of the scores. Overall, GPT-4 with zero-shot rubrics performed the best among all models, and Llama3.1-70B followed closely thereafter. The tuned HHEM-extrinsic model also shows high accuracy with a slight flaw in scoring the sentence-level JsonTG dataset. A slight performance drop can be observed transitioning from the gold hypothesis (\ref{fig:scored_hypo}) to its sentence level (\ref{fig:scored_sent}). GPT-4 scored 6.8\% out of 1393 gold sentences to "1-highly unfaithful", which was mainly due to incomplete information in individual sentences. As a result, several unfaithful occurrences were assigned reason 1 in Table \ref{tab:llm_failed_reason}. Furthermore, JsonTG reference seems to be consistently challenging for all models across all tasks. After inspecting the LLM provided reasoning, we found that this is because our adjective verification rubrics to capture "adjectives without any basis" contradict the JsonTG generation task, which allows LLMs to come up with expressive adjectives. For posterity, it is best to adapt the rubrics to a per-use case basis for better accuracy.

LLMs with fewer parameters, such as Llama3.1-8B and Claude3-Haiku, seem to struggle to understand the rubrics. Llama3.1-8B tends to be overly strict in that it scored all tasks at "3-slightly unfaithful" and below, indicating the hypothesis is unfaithful. This seems to be a result of Llama3.1-8B's following the instructions too literally. In addition, Llama3.1-8B seems to have confused "factual consistency" with "granularity". The example LLM reasoning can be found in rows 2 and 3 in Table \ref{tab:llm_failed_reason}. In contrast, Haiku tends to be extremely lenient. It failed consistently on both hallucination tasks while scoring almost perfectly on the gold hypothesis. It appears that Haiku was able to capture the key information in the hypothesis but could not make the association to the contradictory part in the reference. Row 4 in Table \ref{tab:llm_failed_reason} illustrated an example in which Haiku incorrectly scored an intrinsic hypothesis as "5-highly faithful".

It is interesting that the above observation is not reflected in the overall residual plot in Figure \ref{fig:scored_all}, as the scores were smoothed out by half of the tasks with good performance. However, it should be noted that although the models' strictness and leniency are consistent across different tasks, in reality, it would be difficult to integrate them into the production pipeline due to the unpredictability brought about by their scoring bias.

Unlike LLMs, HHEM models do not experience significant performance degradation from full-length to sentence-level hypothesis. This is likely because the base Deberta model was originally trained for NLI, which uses a slightly fuzzy definition for "textual entailment" that a premise entails a hypothesis given a very probable inference can be made from the premise \citep{dagan2005pascal}. Thus, a piece of a sentence from the hypothesis would receive a higher score during the prediction as long as the encoded information can be inferred from the source. It should be noted that HHEM models are naturally prone to input token size, and evaluating on sentence level could improve efficiency by inducing fewer segments on the reference. 

HHEM models are also less sensitive toward extrinsic hallucinations as compared to intrinsic hallucinations. Fine-tuning on extrinsic synthetic JSON data alone helped the HHEM-extrinsic model to achieve better performance on almost all tasks. However, during fine-tuning, we noticed that it was rather easy to over-tune the HHEM model and made it less tolerant to any type of creative content. We first tried tuning with 400 each gold and hallucination examples for 5 epochs, which resulted in overly strict scores. By reducing the number of epochs to 3, we were able to achieve the performance as shown in Figure \ref{fig:scored_result1}. We believe that if the model is trained on more diverse synthetic examples with a carefully crafted dataset, the results would be more stable than what we have shown here.

\subsection{Scoring Progression}
Figures \ref{fig:scored_result2} illustrate the progression trends of the scores with respect to the percentage of hallucination. Table \ref{tab:step_sensitivity} further shows the sensitivity of each model towards changes in amount of hallucination. To recap, the percentage-based hallucination dataset was created by randomly replacing the N-numbers of gold sentences with hallucinating ones. Each hallucinating sentence contains a subtle change, as demonstrated in Table \ref{tab:hallucination_examples}. Consequently, the hallucination percentage is, in fact, the percentage of sentences containing unfaithful information. A gold sample is considered to contain 0\% hallucination while a sample with only hallucinating sentences is considered to have 100\% hallucination. In this iteration, we also tested the LLM scoring capability when the "reasoning" component was removed from its response. The score-only model's performance is suffixed with "SO".

\begin{table*}[htb!]
    \begin{tabular}{c|ccc|ccc} \hline
         \textbf{hallucination} & \textbf{0\%} & \textbf{100\%} & \textbf{delta per} & \textbf{0\%} & \textbf{100\%} & \textbf{delta per} \\ 
         \textbf{percentage} & \textbf{intrinsic} & \textbf{intrinsic} & \textbf{step (20\%)} & \textbf{extrinsic} & \textbf{extrinsic} & \textbf{step (20\%)} \\ \hline
        GPT-4 & 4.90 & 1.06 & \textbf{-0.77} & 4.90 & 1.29 & -0.72 \\ 
        Llama3.1-70B & 4.92 & 1.12 & -0.76 & 4.92 & 1.27 & \textbf{-0.73} \\ 
        Llama3.1-70B SO & 4.62 & 1.12 & -0.70 & 4.62 & 1.22 & -0.68 \\
        GPT-4o-mini SO & 4.78 & 1.39 & -0.68 & 4.78 & 2.55 & -0.45 \\ 
        GPT-4o-mini & 4.35 & 1.00 & -0.67 & 4.35 & 1.55 & -0.56 \\ 
        GPT-4 SO & 4.84 & 1.57 & -0.65 & 4.84 & 2.16 & -0.54 \\
        HHEM-extrinsic & 4.16 & 1.15 & -0.60 & 4.16 & 1.08 & -0.61 \\ 
        HHEM-intrinsic & 3.83 & 1.10 & -0.55 & 3.83 & 1.96 & -0.38 \\ 
        HHEM & 4.25 & 1.55 & -0.54 & 4.25 & 3.27 & -0.33 \\ 
        Llama3.1-8B SO & 3.06 & 1.08 & -0.40 & 3.06 & 1.20 & -0.37 \\
        Llama3.1-8B & 2.62 & 1.07 & -0.31 &  2.61 & 1.96 & -0.28 \\ 
        Claude3-Haiku SO & 4.61 & 3.67 & -0.19 & 4.61 & 3.82 & -0.16 \\
        Claude3-Haiku & 5.00 & 4.65 & -0.07 & 5.00 & 4.35 & -0.13 \\ \hline
    \end{tabular}
    \caption[]
    {\small \textbf{Model sensitivity vs. hallucination percentage.} The larger the change (delta per step) is, the more sensitive the model is towards the amount of hallucination. In other words, GPT-4 is the best at detecting degrees of intrinsic hallucination, whereas Llama3.1-70B is the best at differencing levels of extrinsic hallucination.}
    \label{tab:step_sensitivity}
\end{table*}

As shown in Figures \ref{fig:scored_mixed_intr} - \ref{fig:scored_mixed_extr} for both intrinsic and extrinsic hallucinations, the faithfulness score decreases accordingly as the percentage of hallucination increases from 0\% to 100\%. This trend is consistent even for lower-performing LLMs such as Llama3-8B and Claude3-Haiku. As mentioned earlier, all models tend to perform better in intrinsic hallucinations than in extrinsic ones. We can potentially explain this observation further by comparing the changes in the score from a lower to a higher percentage of hallucinations. Taking GPT-4 as an example, in Figure \ref{fig:scored_mixed_intr}, when the intrinsic hallucination percentage increases from 0\% -> 20\% -> 40\%, the score decreases from 4.90 -> 3.23 -> 2.26; whereas for extrinsic hallucination, the score decreases in a more gradual manner, namely 4.90 -> 4.13 -> 3.15, with every step taken. GPT-4's sensitivity to identify the initial 20\% of intrinsic hallucination is roughly mapped to detecting 40\% of extrinsic hallucination. Tracing back to our rubrics, although "avoiding speculative or creative content generation" was included in the guidelines, the score assignment was lenient toward extrinsic hallucination. We designed the "3-slightly unfaithful" to allow unverifiable information to be present without negatively impacting the reader's experience. Tightening the rubrics could help increase the sensitivity in detecting extrinsic hallucinations. However, one should proceed with caution and support the rubric's development by using similar experiments to ensure that the LLMs maintain a good tolerance between unfaithful and creative content.

We also quantified the scores without reasoning by simply removing the instruction to justify the reason for the grading. Overall, in Figures \ref{fig:scored_mixed_intr} - \ref{fig:scored_mixed_extr}, all LLMs demonstrated various levels of performance degradation. For both intrinsic and extrinsic hallucinations, scoring without reasoning made the LLM more lenient toward unfaithful content. As the percentage of hallucinations increases, the decreases in the score converge much slower.  The overall decrease in the "delta per step" in Table \ref{tab:step_sensitivity} further justifies that models are less sensitive to extrinsic hallucinations compared to intrinsic ones. Another interesting observation is that the strictness of Llama3.1-8B and the leniency of Claude3-Haiku have both relaxed slightly without reasoning, which caused a performance "improvement". We argue that this is because the reasoning is the key to keeping the model "sharp", the "improvement" is really just a form of unresponsiveness when the "thinking" step is removed from the process.

\begin{table*}[htb!]
    \centering
    \begin{tabular}[width=\textwidth]{p{0.11\textwidth}p{0.095\textwidth}p{0.095\textwidth}p{0.095\textwidth}p{0.095\textwidth}p{0.095\textwidth}p{0.095\textwidth}} \\ \hline
        \textbf{} & \textbf{GPT-4} & \textbf{GPT-4o mini} & \textbf{Llama3.1 70B} & \textbf{Llama3.1 8B} & \textbf{Claude3 Haiku} & \textbf{HHEM models} \\ \hline
        \textbf{reasoning} & 6.82s & 1.08s & 1.85s & 2.02s & 1.94s & N/A \\ \hline
        \textbf{score only} & 1.54s & 0.41s & 1.28s & 1.99s & 2.15s & 0.007s \\ \hline
        \textbf{out tokens} & 117.08 & 109.18 & 24.69 & 121.67 & 95.90 & N/A \\ \hline
        \textbf{price(\$)} & 0.0452 & 0.0003 & 0.0013 & 0.0003 & 0.0004 & N/A \\ \hline
    \end{tabular}
    \caption{\textbf{Utilization per reference \& hypothesis pair in terms of latency in seconds and price.} The number of output tokens containing both reasoning and score was estimated by a \textit{cl\_100k} tokenizer for consistency. Score-only evaluations output 6 tokens consistently. The latency of HHEM models was estimated with batch size 32 on an AWS G5 instance. The LLMs are hosted through the same proxy. It should be noted that LLM's latency can be affected by not only the number of generation tokens but also the overall traffic load.} 
    \label{tab:latency}
\end{table*}

\subsection{Latency \& Usage}
Table \ref{tab:latency} illustrates the latency and price of each model. All LLMs were tested through HTTP endpoints hosted through the same proxy service. The average time in seconds, the number of output tokens and the prices were calculated in a sequential API call setting. By nature, smaller encoder models are inherently faster than decoder-based LLMs; thus we only computed a batch-inference latency for HHEM models run on an AWS G5 instance. The pricing referred to OpenAI \footnote{https://openai.com/api/pricing/} and AWS Bedrock \footnote{https://aws.amazon.com/bedrock/pricing/} standards. While GPT-4 yields the best accuracy, it is also the most costly model in both latency and price. On the other hand, although Llama3.1-70B seems to have balanced out the accuracy vs. latency, Llama3.1-70B provided reasoning can be barely useful to a human auditor due to little granularity. It should be noted that one could customize the prompt instruction to adjust the level of granularity to achieve the desired balance. For real-world usage, we recommend combining a fast model with acceptable performance as a first-pass evaluation; then apply more powerful LLMs with detailed explanation to assist human audits.

\section{Conclusion}
\label{sec:conclusion}

We proposed a rubric-based template to use LLMs to score faithfulness in NLG. We evaluated 5 popular LLMs on 4 travel-domain indutry dataset. The results indicate that in a zero-shot setting LLMs, e.g. GPT-4, outperform NLI models, e.g. HHEM, on capturing both intrinsic and extrinsic hallucinations. HHEM model with fine-tuning on synthetic data can outperform LLMs in domain-specific evaluation given carefully crafted training data. Lastly, we introduced a heuristic to quantify the faithfulness score over degrees of hallucination percentage, which helps to visualize the model's sensitivity. Our future work includes exploring the score association between different LLMs of the same family and tuning the rubrics to adjust the scoring sensitivity. 

\section{Limitations}
\label{sec:limitation}

\begin{enumerate}
    \item Due to resource limitations, we focused on testing only travel-domain dataset, but did not include popular open source faithfulness dataset. We justify that real-world datasets are often more challenging than research domain datasets, as there exists more noise in real-world data. We believe that our approach can be generalized to other domains. 
    \item Due to security concerns, the proxy service used for LLM testing contains enterprise guardrails which moderate incoming and outgoing data. Thus, the latency reported in Table \ref{tab:latency} would be higher than directly calling OpenAI and Bedrock APIs.
    \item Due to resource limitations, we were unable to evaluate a larger model from the Anthropics family, such as the Claude3.5-Sonnet \citep{anthropic2024claude3_5}. It was not our intention to make an unfair comparison using only a smaller Claude model.
    \item Due to privacy concerns, we are in the process of legal review to determine which sections of the dataset can be made public.

\end{enumerate}

\section{Acknowledgment}
\label{sec:Acknowledgment}

This work was conducted by Data and AI, Expedia Group Inc. The authors would like to thank the anonymous reviewers for their detailed, kind and constructive reviews; Anirudh Kamalapuram Muralidhar for the initial work, Adam Johns and the team for the related work on the LLM rubrics; Magdalena Labori and Ananth Raj GV for the valuable feedback; Saman Enayati, Zoe Yang, and Mani Najmabadi for their supports; and David Cohen, Rose James, and Christina Riggs for legal guidance.

\newpage
\bibliography{999_bib}


\appendix

\section{Appendix: Additional Data and Results}

\subsection{Prompt Templates}
\label{app:prompt_templates}
All the prompts in this paper were experimented with a temperature set to 0 for consistency. 

Table \ref{tab:appendix_prompt_grading} contains the prompt used for LLM faithfulness grading. For score-only grading, we simply removed the "reasoning" field from the output format. During our experiments, without enforcing controlled decoding to ensure consistent JSON output, we noticed that only GPT-4 and GPT-4o were able to always respond with a stringified JSON. Both Llama3.1-8B and Claude-Haiku tend to add heading and tailing texts and sometimes include unhashed double quotes, which resulted in a JSON decoding challenge. For open source models which can be served by VLLM \citep{kwon2023efficient}, we recommend using VLLM's controlled decoding strategy to formulate the structured output. 

Table \ref{tab:appendix_prompt_intrinsic} illustrates prompts used for generating the synthetic intrinsic and extrinsic hallucination data. Each prompt follows the "Template" format as noted in row 1. Depending on the use case and the nature of the dataset, we customize the case-by-case instruction for the best results, as indicated in rows 2-7. It should be noted that a general prompt for generating synthetic data could work; however, the data generated might not be tricky enough, so that the changes are subtle. For example, when our source data contain the following ground truth: "3-bedroom" and "2-bathroom". There is a chance that an LLM confuses the numeric values and writes "2 bedrooms and 2 bathrooms" in the generation. By including explicit details in the instructions, we were able to mimic similar cases.

\subsection{Additional Analysis on Synthetic Data}
Table \ref{tab:additional_intr_examples} demonstrates several tricky intrinsic examples in the synthetic dataset. Previous studies have shown that language models tend to struggle to follow instructions when there are negations present in the prompts or generations \citep{truong2023language, varshney2024investigating}. Although a large portion of negations similar to row 1 were scored correctly, we did see that GPT-4 consistently fail on cases like row 2. In this case, the gold summary was, in fact, extracted word by word from the source conversation: \textit{"If your original card charge is still processing, it will be dropped automatically."} The word "not" was obviously ignored every time when evaluating the intrinsic sentence against the source. Rows 3 and 4 are good examples of near-realistic synthetic data. Especially in row 4 where the information is almost correct, but not 100\% accurate. 

Table \ref{tab:additional_extr_examples} presents a few examples of extrinsic hallucinations generated by GPT-4o. We found some of the synthetic sentences extremely humorous, especially when the knowledge is added in an analogy form which potentially triggers a punchline. While these might be too "easy" to capture when served as extrinsic hallucinations, we thought it is worth pointing out this discovery, as it may be useful to humor research with LLMs.

\subsection{Additional Results}
\label{app:additional_results}

We also experimented on Qwen-7B \& 72B \cite{yang2024qwen2}, but did not include the results in the main paper because Qwen-7B occasionally (about 10\% of the times) responded with Chinese reasoning, which caused difficulties in comparison and analysis. Furthermore, we quickly explored fine-tuning HHEM with intrinsic and extrinsic combined synthetic data; however, the initial results indicate that the HHEM-combined was a little too strict in scoring gold data. This was likely due to the hallucination samples being double in size than the gold samples after combining the datasets. Due to resource limitation, we did not further adjust the training data to achieve the desired results. In Table \ref{tab:residuals_scored_all}, we show the residual values by task for Qwen, HHEM-combined, and all models presented in the main paper as a reference.

\subsection{Additional Discussion on Evaluation Subjectivity}
One may argue that a rubric-based approach is subject to the particular use case, which can be challenging across different contexts. This concern stands for not only the method proposed in this paper, but also other polarity-based faithfulness classification models. Similarly to any traditional machine learning methods, the definition or gold data is defined by human annotators. A model can also be overfitting to a particular use case if the gold labels are very domain-specific. For example, in opinion mining, "Object A is very fast" can indicate a positive opinion if "Object A" is a calculator, but it can also indicate a negative opinion if "Object A" is a car running at full speed in a residential area with young children. Now this example might seem a bit extreme; the core is that while consistency is indeed important, subjectivity is not 100\% bad and it can sometimes help domain-specific adaptions. 

On the other hand, in our case, the subjectivity only resides in the quantifiable metrics, but not in the domain itself (e.g. travel). We argue that quantifiable metrics should be viewed as a separate component as the domain. We view quantifiable metrics as downstream tasks which can be combined with flexibilities. In Table \ref{tab:appendix_prompt_grading}, we attached the prompt used to score summaries during the experiments. The prompt is domain independent, which can be applied to any summarization tasks. Because some quantifiable metrics such as adjective usage and tone of voice can be subjective depending on the use case, having the ability for the user to tailor the rubrics will best help the LLM judge fit to that particular use case. For example, it is possible in stylized text generation that having LLMs add descriptive adjectives and verbs is desired in the generation, in which case the user can adjust the “adjective consistency” requirements in the rubrics to fit that use case. We intended to introduce an adaptable solution, and similar to any other ML models, small adjustment might still be required when adapting to granular details.

Finally, on concerns over generalization. We believe that in the current era of LLMs, generalization of GenAI downstream tasks should not be strictly defined as a "universal model that powers all use cases". Rather, generalization should aim for a "framework" or "template" that can be adopted easily by other domains. With a newer and better model evolving at a monthly basis, we believe the prompt templates and experiment designs we proposed in this paper can be generalized to changes.

\subsection{Additional Discussion on Organic vs. Synthetic Hallucinations}
Due to the lack of organic hallucinations in applications, we focused on synthetic hallucinations generated based on real-world observations. One may raise the concern that it is risky to draw conclusions on synthetic hallucinations over organic ones, as the distribution may not be fully modeled. We agree that organic hallucinations are more valuable and that in ideal scenarios, we should develop such faithfulness models using organic data as much as we can. However, we believe that in the real-world case-by-case scenario, it is rather difficult to obtain a large portion of organic hallucinations for development. We believe that for a well-defined system with a trained developer, even the initial development stage would suffer from insufficient negative examples. However, this does not mean that the evaluation should be omitted. It is rather common for developers to generate synthetic test cases (or edge cases) for testing. Thus, we tailored the synthetic data generation for each use case to tackle tricky and subtle inconsistencies. 

We believe that the conclusion on the experiments can still be generalized because 1) the real-world dataset often contains more noise and is more complex in nature than the academic gold dataset. For example, ConvoTS datasets contain many turns of redundant user interactions due to the limitation of the virtual assistant. 2) A portion of the synthetic changes are tricky enough to mimic organic hallucinations (see examples in Table \ref{tab:additional_intr_examples}). Furthermore, even the best LLM judge (GPT-4) failed in some of those tricky cases.



\begin{table*}[htb!] \bitsmall
    \begin{tabular}{p{0.95\linewidth}} \hline
        \textit{\textbf{System Prompt}}  \\ \hline
            Your task is to evaluate the accuracy of a summarized text generated from a source document by comparing it to the original document. Rely solely on the provided source document for verification. The content should be precise and only include verifiable information that is explicitly stated in the source document, so do not make any assumptions or derive any thoughts. Pay close attention to the following aspects of accuracy: \\
            \\
            - Factual Consistency: Check if the summarized text accurately represents factual information from the source document. If the source mentions singular versus plural items (e.g., one restaurant versus multiple restaurants), the summary should reflect this accurately. \\
            - Adjective Usage: Examine the use of adjectives in the description and verify if they are supported by the source document. It's acceptable for the description to paraphrase or use synonyms, but if an adjective lacks any basis in the source document, it should be considered inaccurate. \\
            - Avoid Speculation: Do not engage in speculative or creative content generation. The generated summary should strictly adhere to the factual content of the source document. \\
            - Alignment with Source: The summary should closely align with the source document's tone, language, and content, without introducing any subjective interpretations. \\
            \\
            Please refer to the 'SUMMARY SNIPPET' for the summarized content and 'SOURCE SNIPPET' for the original source text, and then assign a relevance score based on the following grading rubric: \\
            \\
            \# GRADING RUBRIC \\
            Score Accuracy according to the following grading rubric: \\
            - 5: Highly Accurate - All information in the generated content can be verified in the source document. \\
            - 4: Very Accurate - Most information in the generated content can be verified in the source document, with one minor item that wouldn't negatively impact the reader's experience. \\
            - 3: Moderately Accurate - More than one piece of information in the generated content cannot be verified in the source document, but none of these inaccuracies would negatively impact the reader's experience. \\
            - 2: Somewhat Accurate - One or more pieces of information in the generated content are factually inaccurate and cannot be verified in the source document, some, or all of which would negatively impact the reader's experience. \\
            - 1: Highly Inaccurate - Most or all of the information in the generated content is inaccurate, cannot be verified in the source document, and would negatively impact the reader's experience. \\
            \\
            Use this rubric to assess the accuracy of the summarized content compared to the source document in any domain. \\
            \\ \hline

        \textit{\textbf{User Prompt}} \\ \hline
            \# SUMMARY SNIPPET \\
            \{\{ text \}\} \\
            \\
            \# SOURCE SNIPPET \\
            \{\{ source \}\} \\
            \\
            \# OUTPUT \\
            Please provide the scores in a stringified JSON format with two keys: one describing the reasoning for the grading, and one containing the score based on the grading rubric. An example format is as follows, do not include heading or tailing texts: \\
            \{"reasoning": "A detailed explanation for why the score was chosen", "score": 1\} \\ \hline

    \end{tabular}
    \caption{Example prompt for faithfulness scoring of summarization.}
    \label{tab:appendix_prompt_grading}
\end{table*}

\begin{table*}[htb!]
    \begin{tabular}{p{0.01\linewidth}p{0.1\linewidth}p{0.81\linewidth}} \hline

         & \textbf{Type} & \textbf{User Prompt} \\ \hline

        1 & Template & Case-by-case instruction. \\
          & & ===== \\
          & & SENTENCE: \\
          & & \{\{sentencized\}\} \\
          & & ===== \\
          & & REWRITE: \\ \hline

        2 & ConvoAS ConvoTS Intrinsic & Based on the following SENTENCE, write a similar sentence with refuting facts. This should be done through one of the following: changing the value of either important names, named entities, proper nouns, dates, or other numeric values; or through manipulating contradictory information. \\ \hline

        3 & JsonTG Intrinsic & Based on the following SENTENCE, write a similar sentence with refuting facts. This should be done through changing one of the following: a numeric value, the property location, property type, city or name entity name, or facts related to the amenities. The rewrite should contain contradictory information. Make the change subtle. \\ \hline 

        4 &ReviewTS Intrinsic & Based on the following SENTENCE, write a similar sentence with a refuting fact. Try to make the change subtle. \\ \hline
    
        5 & ConvoAS ConvoTS Extrinsic & Expand on the following SENTENCE by adding 1 piece of new fact. This should be done through adding one of the following: unrelated knowledge, topics, named entities, events, or other information that is not directly related to the original SENTENCE. Try to make the change subtle. \\ \hline

        6 & JsonTG Extrinsic & Expand on the following SENTENCE by adding 1 piece of new fact. This can be done by adding one of the following: unrelated knowledge, points-of-interest, named entities, near by events, or other exhilarated travel information that is not directly related to the original SENTENCE. Try to make the change subtle. \\ \hline

        7 & ReviewTS & Expand on the following SENTENCE by adding new facts. This should be done through adding unrelated knowledge that is not directly related to the original SENTENCE. Try to make the change subtle. \\ \hline

    \end{tabular}
    \caption{Example prompts for generating synthetic intrinsic hallucinations}
    \label{tab:appendix_prompt_intrinsic}
\end{table*}

\begin{table*}[ht!]
    \begin{tabular}[width=\textwidth]{p{0.01\linewidth}p{0.1\linewidth}p{0.1\linewidth}p{0.56\linewidth} p{0.1\linewidth}}\\ \hline
         & \textbf{Dataset} & \textbf{Type} & \textbf{Sentence} & \textbf{Scored} \\ \hline
        1 & ConvoAS & Gold & [NAME] informed [NAME] that the travel agency \textbf{does not offer} an installment payment option. & 5-highly faithful \\
         & & Intrinsic & [NAME] informed [NAME] that the travel agency \textbf{does offer} an installment payment option. & 1-highly unfaithful \\ \hline

        2 & ConvoTS & Gold & If the original card charge is still processing, it \textbf{will be} dropped automatically. & 5-highly faithful\\
         & & Intrinsic & If the original card charge is still processing, it \textbf{will not be} dropped automatically. & 5-highly faithful \\ \hline

        3 & JsonTG & Gold & As the evening settles in, gather around the cozy \textbf{fireplace} and enjoy quality time together. & 5-highly faithful \\
         & & Intrinsic & As the evening settles in, gather around the spacious \textbf{outdoor fire pit} and enjoy quality time together. & 2-very \; unfaithful \\ \hline

        4 & ReviewTS & Gold & The saltwater pool was relaxing and comfortable, and it was in \textbf{a sound proof room}. & 5-highly faithful\\
         & & Intrinsic & The saltwater pool was relaxing and comfortable, and it was in \textbf{a mostly sound proof room}. & 4-very faithful\\ \hline

    \end{tabular}
    \caption{Examples of tricky intrinsic hallucination examples from each dataset}
    \label{tab:additional_intr_examples}
\end{table*}

\begin{table*}[ht!]
    \begin{tabular}[width=\textwidth]{p{0.01\linewidth}p{0.1\linewidth}p{0.15\linewidth}p{0.65\linewidth}}\\ \hline
         & \textbf{Dataset} & \textbf{Type} & \textbf{Sentence} \\ \hline

        1 & ConvoAS & Gold, \textbf{Extrinsic} & The customer was satisfied with the explanation and did not have any further requests, \textbf{much like how Jane Goodall felt content after her groundbreaking research on chimpanzees.} \\ \hline

        2 & ConvoTS & Gold, \textbf{Extrinsic} & The VA changed the booking dates and room type, provided the new total cost, and attempted to check the refund status but encountered an issue, \textbf{much like the technical difficulties experienced during the Apollo 13 mission.} \\ \hline

        3 & ConvoTS & Gold, \textbf{Extrinsic} & If the traveler cancels the insurance policy, they will not be able to rebook it for the current itinerary and will not be able to file a claim against the policy, \textbf{much like how concert tickets for Taylor Swift's Eras Tour are non-refundable once purchased.} \\ \hline

        4 & ReviewTS & Gold, \textbf{Extrinsic} & The pool was cold and appeared stagnant, \textbf{much like the ancient Roman baths that were once a hub of social activity, where people would gather not only to bathe but also to discuss politics and philosophy.} \\ \hline

    \end{tabular}
    \caption{Examples of humorous extrinsic hallucination examples from each dataset. The first half (non-bold) is the gold sentence, and the latter half (\textbf{bold}) is the extrinsic knowledge added by GPT-4o. All above were scored correctly as "1-highly unfaithful"}
    \label{tab:additional_extr_examples}
\end{table*}

\begin{table*}[b!]
    \begin{tabular}{p{0.2\linewidth}p{0.13\linewidth}p{0.13\linewidth}p{0.13\linewidth}p{0.13\linewidth}p{0.13\linewidth}} \hline
         & \textbf{Gold} \textbf{}& \textbf{Gold} & \textbf{Intrinsic} & \textbf{Extrinsic} & \textbf{All Tasks} \\ 
         & \textbf{full-length} & \textbf{sent-level} & \textbf{sent-level} & \textbf{sent-level} & \textbf{(mean)} \\ \hline
        GPT-4 & 0.23 & 0.77 & 0.43 & 0.89 & 0.66 \\ 
        GPT-4o-mini & 1.23 & 1.34 & 0.23 & 1.12 & 0.92 \\ 
        Llama3.1-70B & 0.84 & 1.22 & 0.49 & 0.91 & 0.86 \\ 
        Llama3.1-8B & 3.01 & 3.48 & 0.10 & 0.12 & 1.36 \\ 
        Claude3-Haiku & 0.15 & 0.37 & 2.24 & 3.14 & 1.79 \\ 
        Qwen2.5-72B & 0.40 & 0.44 & 1.04 & 2.09 & 1.13 \\ 
        Qwen2.5-7B & 2.85 & 3.22 & 0.64 & 0.78 & 1.64\\ 
        HHEM & 0.53 & 0.57 & 1.02 & 2.24 & 1.22 \\ 
        HHEM-intrinsic & 0.65 & 0.79 & 0.32 & 1.02 & 0.71 \\ 
        HHEM-extrinsic & 0.49 & 0.90 & 0.59 & 0.21 & 0.56 \\ 
        HHEM-combined & 0.98 & 0.83 & 0.34 & 0.24 & 0.51 \\ 
        Baseline & 2.24 & 2.64 & 1.25 & 1.30 & 1.77 \\ \hline
        Number of pairs & 324 & 1393 & 1393 & 1393 & Total 4503 \\ 
        Expected score & 5 & 5 & 1 & 1 & N/A \\\hline
    \end{tabular}
    \caption{Overall residual (absolute error) by task on all models tested. The numeric values match with the visualizations presented in Figures \ref{fig:scored_result1}. For models reported in the main paper, GPT-4 and HHEM-extrinsic performed the best overall. Although Llama3.1-8B and Claude3-Haiku showed very low error rate on half of the tasks, the models are either too strict or too lenient to cover all cases. Additional models, Qwen 2.5 7B \& 72B and HHEM-combined are included for reference.}
    \label{tab:residuals_scored_all}
\end{table*}

\end{document}